\documentclass[10pt,twocolumn,letterpaper]{article}

\usepackage{iccv}
\usepackage{graphicx}
\usepackage{times}
\usepackage{epsfig}
\usepackage{amsmath}
\usepackage{amssymb}
\usepackage{here}
\usepackage{caption}
\usepackage{comment}
\usepackage{scalefnt}
\usepackage{ulem}

\def \figref  #1{Figure~\ref{#1}}

\def \equref  #1{Equation~\eqref{#1}}

\usepackage[breaklinks=true,bookmarks=false]{hyperref}

\iccvfinalcopy 

\def\iccvPaperID{3808} 

\ificcvfinal\fi

\begin{document}

\title{Image Generation From Small Datasets via Batch Statistics Adaptation}

\author{
    Atsuhiro Noguchi${}^\text{1}$ and Tatsuya Harada${}^\text{1,2}$\\\\
    ${}^\text{1}$The University of Tokyo, ${}^\text{2}$RIKEN
}

\twocolumn[{%
\renewcommand\twocolumn[1][]{#1}%
\maketitle
\begin{center}
\centering
\includegraphics[width=0.8\textwidth]{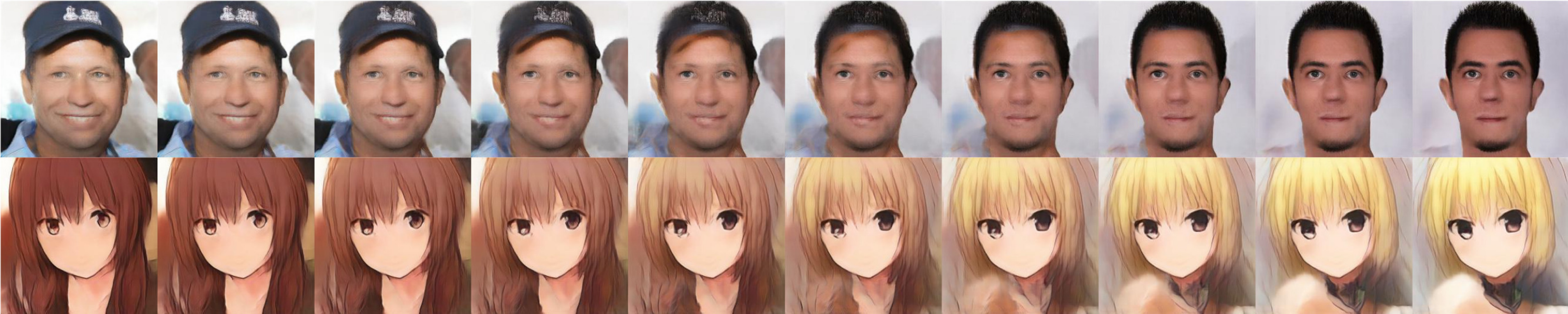}
\captionof{figure}{Interpolation results generated by BigGAN adapted to {\bf only 25} human and anime face images. The left- and right-most images are the generated images that correspond to training samples, and the other images are generated by linearly changing the latent vector. In spite of the small amount of training data, our method achieves a smooth interpolation.}
\label{interpolation}
\end{center}%
}]

\maketitle
\ificcvfinal\thispagestyle{empty}\fi

\begin{abstract}
Thanks to the recent development of deep generative models, it is becoming possible to generate high-quality images with both fidelity and diversity. However, the training of such generative models requires a large dataset.
To reduce the amount of data required, we propose a new method for transferring prior knowledge of the pre-trained generator, which is trained with a large dataset, to a small dataset in a different domain. Using such prior knowledge, the model can generate images leveraging some common sense that cannot be acquired from a small dataset.
In this work, we propose a novel method focusing on the parameters for batch statistics, scale and shift, of the hidden layers in the generator.
By training only these parameters in a supervised manner, we achieved stable training of the generator, and our method can generate higher quality images compared to previous methods without collapsing, even when the dataset is small ($\sim$100). Our results show that the diversity of the filters acquired in the pre-trained generator is important for the performance on the target domain.
Our method makes it possible to add a new class or domain to a pre-trained generator without disturbing the performance on the original domain. Code is available at \scalebox{0.77}[0.9]{\textcolor[rgb]{1.0, 0.0, 1.0}{\url{github.com/nogu-atsu/small-dataset-image-generation}}}

\end{abstract}
\vspace{-3mm}
\section{Introduction}
In recent years, image generation using deep generative models has rapidly developed, and some state-of-the-art methods can generate images that cannot be distinguished from real data \cite{vae, gan, dcgan, pixelrnn,pggan, sngan, sagan,introvae, highAR}.
Typical generative models include Variational Auto-Encoders (VAEs) \cite{vae, vae2}, Generative Adversarial Networks (GANs) \cite{gan}, and Auto-Regressive (AR) models \cite{pixelrnn}. Although these methods can generate novel images that are not included in the training dataset, these generative models have many parameters. For instance, Spectral Normalization GAN with a projection discriminator (SNGAN projection) \cite{sngan, projection} for 128 $\times$ 128 sized images has 90 M trainable parameters. Accordingly, we need a large dataset to train such a large network without overfitting.
In general, a large dataset ($\sim$10,000) is required to train generative models \cite{dcgan, pggan, sngan, projection}. However, constructing such a huge dataset requires significant effort, and a conventional generative model cannot be applied to a domain in which collecting sufficient data is difficult. Therefore, training a generative model from a small dataset is crucial. Moreover, because generative models can learn the data distribution, they have an advantage not only in generating images but also in improving the performance of classification and abnormality detection models through semi-supervised learning \cite{cvae, gan_anno}. If we can train a generator from a small dataset, the performance of these tasks can be improved by interpolating the training data.

The transfer of prior knowledge is effective to train deep-learning models from a sparsely annotated or small dataset. For a feature
extractor model, transfer learning, in which a feature extractor
model is trained using a large labeled dataset and is transferred to another domain with sparse annotation, has been widely studied \cite{fine, dan, dann, adda, dsn, pixelda}. Training on a target dataset starting from the weights of the pre-trained model is called fine-tuning \cite{fine}. Even when fine-tuning the model with a dataset in a completely different domain from the dataset used to train the pre-trained model, the performance tends to be better than training from scratch. This is because a pre-trained model acquires generally useful weights that cannot be obtained using a small target dataset.
A method for transferring prior knowledge to another dataset has been proposed for generative models as well \cite{transfergan, transferrnn}. It was shown that transferring prior knowledge also improves the performance of generative models.

To adapt prior knowledge, we focus on the scale and shift parameters of batch statistics in the generator. These parameters can be seen to control the active filter in the convolution layer, and it can be stated that updating the scale and shift parameters selects filters which are useful for generating images similar to the target domain. We propose a new transfer method for a generator that updates only the scale and shift parameters in the generator. By updating only these parameters and fixing all kernel parameters in the generator, we can reduce the number of images required to train the generator.
We conducted experiments by applying this method to a very small dataset consisting of less than 100 images and showed that the quality is higher than that of previous methods, and that it is possible to generate images capturing the data semantics.

\section{Related works}
In this paper, we propose a novel method for transferring a pre-trained generative model to realize image generation from a small dataset. We introduce related studies on generative models in subsection \ref{dgm}, transfer learning for generative models in subsection \ref{tf}, and transfer learning that uses scale and shift parameters in subsection \ref{ts}. 

\subsection{Deep generative model\label{dgm}}
Studies on data generation using deep learning techniques have been developed rapidly in recent years, and methods that can generate data such as images and languages have been widely studied \cite{vae, vae2, gan, pixelrnn}. Typical generative models include VAEs \cite{vae, vae2}, GANs \cite{gan}, and AR models \cite{pixelrnn}. VAEs model variational inference and learn to maximize the variational lower bound of likelihood. GANs consist of two networks, a generator and a discriminator. The generator generates data close to the training data, and the discriminator identifies whether the input data are training or generated data.
By training these models in an adversarial manner, the generator becomes able to generate data that are indistinguishable from the training data. AR models express the data distribution as a product of the conditional probabilities, and sequentially generates data. Techniques that can generate consistent high-quality images through any of these methods have recently been developed \cite{introvae, sngan, sagan, biggan, highAR}.

Every model has a large number of trainable parameters, and a large dataset is necessary to prevent overfitting. For example, in SNGAN projection for 128 $\times$ 128 sized images, there are 90 M trainable parameters. With GANs, in particular, the discriminator estimates the distance between the distributions of the real and generated data. Therefore, training a discriminator requires a large dataset sufficient to fill in the distribution of the real data. The training of these models often uses a dataset on the order of more than 10,000 examples.
Because it is extremely time-consuming to construct such a large dataset, it is an important task to reduce the amount of data necessary to train a generative model.
Since a generative model learns the distribution of data, there is an advantage in that it can be used for classification tasks, abnormality detection, and so on through semi-supervised learning \cite{cvae, gan_anno}, and it is expected that the construction of a generative model from a small dataset can contribute to these fields.

\subsection{Transfer learning for generative models\label{tf}}
It is known that transfer learning is effective in improving the performance for sparsely annotated or limited data \cite{fine, dan, dann, adda, dsn, transfergan}. Transfer learning transfers the knowledge acquired through training on a large dataset to another dataset of a different domain for which there are insufficient labels.

A method for transferring generative models learned with a sufficient amount of data to another dataset was developed \cite{transferrnn, transfergan}. In \cite{transferrnn}, the techniques for knowledge transfer for neural language model is proposed, and in \cite{transfergan}, a pre-trained GAN model is fine-tuned to the target domain to transfer the knowledge. These techniques enable the generative model to converge faster and obtain better performance than with normal training. Results suggest that transferring pre-trained knowledge is effective for generative models as well. However, especially in image generation \cite{transfergan}, 1,000 training examples are still necessary.

\subsection{Transfer learning with scale and shift\label{ts}}
There are some transfer learning methods that modulate only scale and shift parameters in the hidden activations \cite{adabn, mtl}.
Adaptive batch normalization \cite{adabn} performs domain adaptation for the segmentation task by replacing the statistics of the source domain with those of the target domain, achieving performance accuracy competitive with other deep-learning based methods. Meta-transfer learning \cite{mtl} performs few-shot learning by updating only the scale and shift parameters, and achieves better performance than when fine-tuning all kernel parameters. These methods show that the scaling and shifting operation is effective for transferring knowledge to feature extractor models.

The question is whether these operations can also transfer generative knowledge in the generator to images that do not appear in the training samples. To confirm the transferability, we investigate the role of the batch statistics and analyze them in the pre-trained model in the next section.

\section{Role of Batch Statistics}
In this research, we use scale and shift parameters to transfer the knowledge acquired in the pre-trained generator. To show the property of the scale and shift parameters, we discuss the role of these parameters from the point of view of filter selection, and analyze how the filters are selected in the pre-trained SNGAN projection \cite{sngan, projection} model.

\subsection{Scale and shift for filter selection\label{filter_sele}}
In this subsection, we provide a brief analysis of our method in terms of filter selection.
A convolution can be seen as a combination of filters that convert a three-dimensional tensor into a scalar. The number of filters is the same as the number of output channels of the convolutional layer.
In this case, applying the scale and shift to the results of the convolution operation is equivalent to the following convolutional operation.
\begin{align}
&conv(x;W) \cdot \gamma + \beta\notag \\
&= conv(x;W \cdot \gamma + \beta)\notag \\
&= conv(x;\{\gamma_1 W_1+\beta_1, ...,\gamma_{c_{out}} W_{c_{out}}+\beta_{c_{out}} \})
\end{align}
Here, $W$ is a four-dimensional tensor representing the weight of the convolution, and $W_i$ represents the $i^{\text{th}}$ filter in the convolution. In addition, $c_{out}$ is the number of output channels for the convolution. This means that changing the scale $\gamma$ is equivalent to changing the activation strength of the filter of each convolution. In addition, changing the shift $\beta$ means changing the activation threshold of the filter. When $\gamma_i$ and $\beta_i$ are large, the corresponding neuron becomes easy to activate, and when $\gamma_i$  and $\beta_i$ are small, it becomes less active. 
We conducted an experiment and confirmed that there is a positive correlation between $\gamma$ and $\beta$ and the activation rate of the filter. The result is given in the supplementary materials. Therefore, it is shown that changing the scale and shift parameters is equal to performing filter selection and controlling the activation in a Convolutional Neural Network (CNN).

\subsection{Analysis of scale and shift in SNGAN\label{scale_shift_analysis}}
In SNGAN projection \cite{sngan, projection}, class conditional batch normalization is used, where different scales and shifts are applied for each class. That is, by using different $\gamma$ and $\beta$ during batch normalization \cite{BN}, the model creates a difference in the distribution for each class. In the previous subsection, we stated that these parameters control the activity of the filter. In this subsection, we discuss how filters are selected for each class in SNGAN trained on ImageNet. Because there is a correlation between the activation of the filter and $\gamma$ and $\beta$, we only have to check the $\gamma$ and $\beta$ acquired for each class.

\begin{figure}[t]
\center
\includegraphics[width=1.\linewidth]{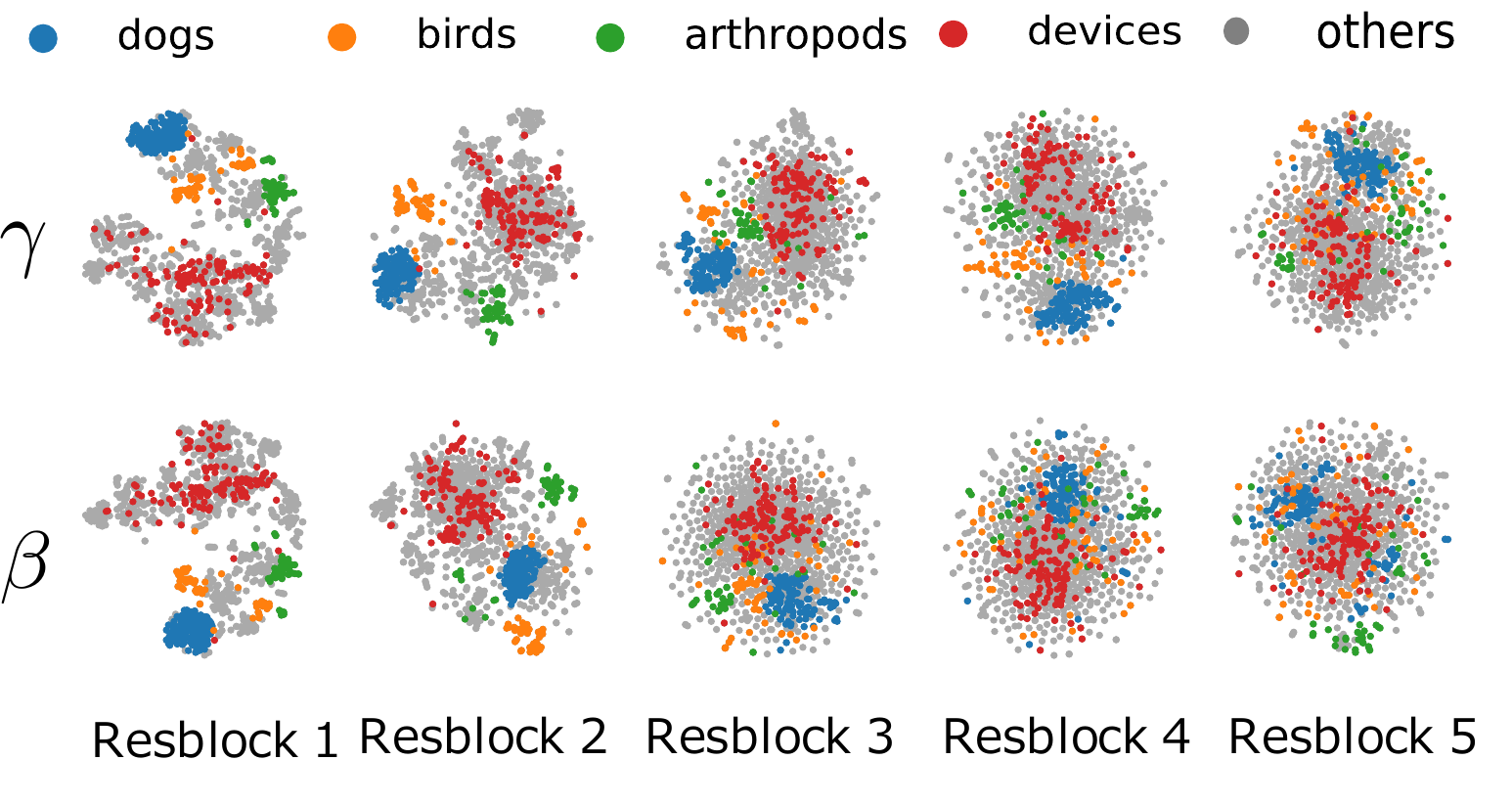}
\vspace{-5mm}
\caption{T-SNE on $\gamma$ and $\beta$ for each layer, where $\gamma$ and $\beta$ are scale and shift parameters respectively.}
\label{tsne_plot}
\end{figure}

For SNGAN trained on ImageNet, we plotted the distribution of $\gamma$ and $\beta$ using T-SNE, as shown in \figref{tsne_plot}. Each point corresponds to each class. We plotted some classes in the categories ``dogs", ``birds", ``arthropods", and ``devices" using different colors, and used WordNet \cite{wordnet} to categorize these classes. Based on this, it turns out that a similar scale and shift are used for semantically similar classes.

This suggests that SNGAN trained using ImageNet acquires various filters, and the model learns the method for selecting useful filters for the generation of each class. If there are sufficiently diverse filters in the trained generator, it seems that there is a possibility for data of different domains to be generated by learning a method for selecting useful filters based on the semantics.

\section{Method}
In this paper, we propose a method for adapting pre-trained generative models to datasets of different domains. Our method only requires a pre-trained generator, and we can leverage any type of generator using a CNN, such as GANs or VAEs. By introducing scale and shift parameters to each hidden activation of the generator, and updating only these parameters, the generative model can be transferred to a small dataset, reducing the number of trainable parameters.

\subsection{Learnable parameters}
To use the prior knowledge obtained, the adapted generator is trained from the weights acquired in the pre-trained generator. However, the number of parameters in a CNN generator is extremely large, and if the available training data size has relatively few samples, the model tends to immediately overfit to the dataset. Therefore, we do not update the kernel parameters of the model in any way, and use scale and shift to control the activation of the filters. This means that updating these parameters may be sufficient for adaptation if there are diverse filters in the generator. We introduce the scale and shift parameters for each channel of the hidden layer distribution of each layer (excluding the final layer) and update only these parameters to conduct an adaptation.
\begin{align}
G^{(l)}_{Adapt} = G^{(l)} \cdot \gamma^{(l)} + \beta^{(l)}
\end{align}
Here, $G^{(l)}$ is the feature representation of the $l^{\text{th}}$ layer of the generator, $G^{(l)}_{Adapt}$ is the feature of the $l^{\text{th}}$ layer after adaptation, $\gamma^{(l)}$ represents the scale parameter of the distribution for adaptation, and $\beta^{(l)}$ represents the shift parameter of the distribution. The initial value of the $\gamma$ element is 1, and the initial value of the $\beta$ element is zero.

Because $\gamma$ and $\beta$ are used in the batch normalization layer, we update these parameters for the scale and shift without adding new statistics parameters. We fix \textit{running\_mean} and \textit{running\_var}  during training. For class conditional batch normalization used in SNGAN projection, different $\gamma$ and $\beta$ values in the batch normalization are used for each class labels.
In this case, we initialize $\gamma$ and $\beta$ to 1 and zero, respectively, and then fine-tune them.

\subsection{Training}
For GANs, the discriminator distinguishes between real images in the dataset and the generated images, and the generator generates realistic images by conducting adversarial training \cite{gan}. However, this method is based on the fact that training data densely fill the distribution, and it suffers from overfitting for small datasets, resulting in unstable training. Therefore, it is desirable to conduct training in a supervised learning framework such as VAEs. However, it is difficult to learn an encoder (such as VAEs) from scratch when only a small dataset is available.

\begin{figure}[t]
\center
\includegraphics[width=.9\linewidth]{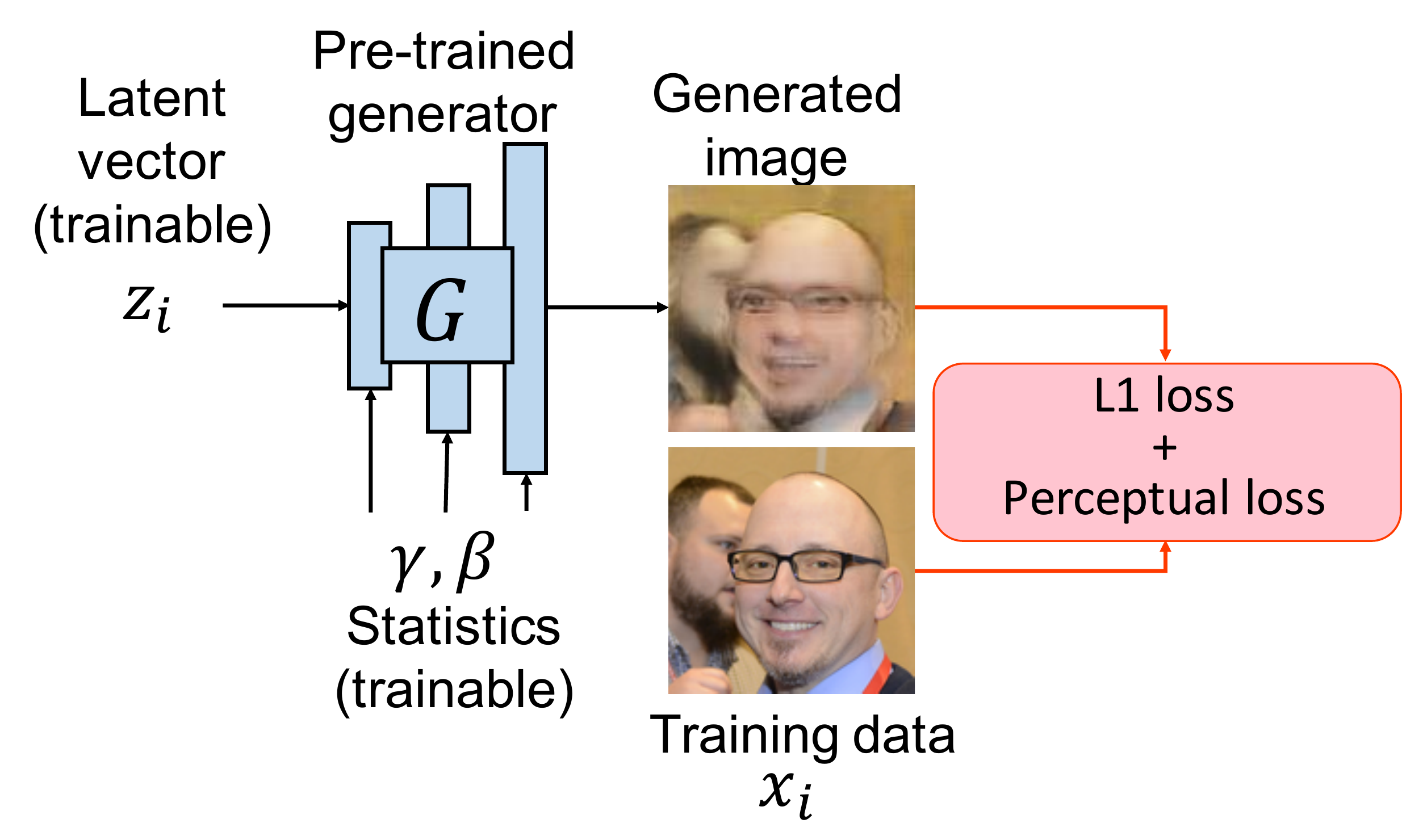}
\vspace{-3mm}
\caption{Proposed pipeline. During training, the scale and shift of the generator and latent variable $z$ are updated to minimize the loss using the L1 and perceptual losses between all training data and the generated data.}
\vspace{-3mm}
\label{pipe}
\end{figure}

To train the generator using supervised learning, the correct target data corresponding to the latent vectors are necessary. Therefore, in this study, we realize supervised learning by simultaneously estimating the latent variable $z$ for all training data such that the generated data are close to the image in the training dataset, which is similar to Generative Latent Optimization \cite{glo}.
The proposed pipeline is shown in \figref{pipe}. During training, loss function $L$ modelled as the distance function to the target image is optimized. $L$ uses the L1 loss, which is the distance at the pixel level, and perceptual loss \cite{perceptual}, which is the distance at the semantic level.
\begin{align}
L =& \sum_i  \frac{1}{c_x h_x w_x} ||x_i - G_{Adapt}(z_i + \epsilon)||_1 \notag \\
&+  \sum_i \sum_{l \in layers} \frac{\lambda_C^l}{c_l h_l w_l}||C^{(l)}(x_i) - C^{(l)}(G_{Adapt}(z_i + \epsilon))||
\notag \\
&+ \lambda_z (\sum_j^k \frac{1}{d_z}\min_i ||z_i - r_j||_2^2 + \sum_i^b \frac{1}{d_z}\min_j ||z_i - r_j||_2^2)\notag \\
&+ \lambda_{\gamma, \beta} \sum_l \frac{1}{d_{\gamma, \beta}^l} (||\gamma_l - 1||_2^2 + ||\beta_l||_2^2)
\label{loss}
\end{align}
Here, $ x_i $ is the $ i^{\text{th}}$ image, and $ z_i $ is the latent vector corresponding to $ x_i $. In addition, $c$, $h$, $w$, and $d$ are the channel, height, width, and dimension of each feature, respectively, $ C ^ {(l)} $ is the feature of the $ l^{\text{th}}$ layer of the trained classifier $C$, $ layers $ are the layers used for perceptual loss, $b$ is batchsize, $ \lambda $ is the coefficient used to determine the balance of each term, and $ r_j \sim \mathcal N (0, I) $ is a vector randomly sampled from the normal distribution. Here, $k$ is a number sufficiently larger than the amount of training data, and $\epsilon $ is a small random vector to avoid local minina. The first and second terms are used to make the generated image close to the training image at the pixel and semantic levels. The third term is used to regularize $z$ as a standard normal distribution. The fourth term prevents overfitting. In this study, we used VGG16 \cite{vgg} for $C$.

Here, $ z $ corresponding to each training data is initialized with a zero vector and is learned via loss minimization using the gradient descent method.

\subsection{Inference}
During inference, by inputting a randomly sampled vector $z$ according to the standard normal distribution to the generator, it is possible to generate images randomly.
However, the generator only learns the relationship between latent vectors and sparse training samples, resulting in the poor performance for $z$ which is far from any training samples. To solve this problem, we sample $z$ from a truncated normal distribution, and this technique is known as the truncation trick \cite{biggan}. The details are described in \ref{generation_small}.

\section{Experiments}
Some experiments were conducted to evaluate the difficulty and possibility of image generation from a small dataset, as well as the stability of the proposed method. We also compared our method to existing studies.
For the generator, we used SNGAN \cite{sngan} and BigGAN \cite{biggan}. Considering the computational costs, we used SNGAN for the comparison experiments. We used 128 $\times$ 128 images for SNGAN and 256 $\times$ 256 images for BigGAN. All experiments in this paper are not class conditional, but our method can be easily extended to be class-conditional by learning BatchNorm-statistics for each class independently, which is similar to SNGAN and BigGAN.

\subsection{Datasets and evaluation metrics}
We used the facial images from FFHQ dataset \cite{stylegan} and anime face dataset\footnote{\textcolor[rgb]{1.0, 0.0, 1.0}{\url{www.gwern.net/Danbooru2018}}} and images of passion flowers from the Oxford 102 flower dataset \cite{flower}. The domains used in this experiment, ``human face", ``anime face", and ``passion flower" are not contained in the classes of ImageNet and can, therefore, be considered different domains to ImageNet. Especially, anime faces never appear in ImageNet.

We employed the commonly-used Fr\'echet inception distance (FID) \cite{fid} for the evaluation of generated images. To effectively evaluate overfitting to training samples, we calculated the distance between 10,000 generated images and randomly sampled 10,000 images from the whole datasets. Though FID is a widely used evaluation metric, it is known that FID is not stable for calculating the distance for small sets of images. Because the flower dataset consists of only 251 images, we also report the more stable Kernel Maximum Mean Discrepancy (KMMD) for all experiments.  We calculate KMMD with a Gaussian kernel between the features in a pre-trained inception network of training and generated images. Lower FID and KMMD indicate better performance.

\subsection{Image generation from a small dataset using comparison methods}
In this section, to confirm the difficulty of training generative models from a small dataset, we conducted experiments on the following seven methods, two of which are previous approaches, and the others are methods similar to our method. 

\noindent{{\bf GAN from scratch}}:
In this method, we trained a GAN from scratch on a small dataset. We used the SNGAN model used in \cite{sngan}.

\noindent{{\bf Transfer GAN} \cite{transfergan}}:
The pre-trained generator and discriminator are fine-tuned on a small dataset. 

\noindent{{\bf Transfer GAN (scale and shift)}}:
This method is similar to our method, but does not apply supervised training, and instead uses unsupervised training with the discriminator. In this method, only the scale and shift parameters in the generator and discriminator are updated to prevent overfitting. 
In SNGAN, spectral normalization on the discriminator guarantees the discriminator as a Lipschitz function \cite{sngan}. Therefore, to maintain this constraint, we also applied spectral normalization to the scale parameters. That is, the scale was divided by the maximum value of the scale.

\noindent{{\bf Encoder-generator network}}:
Our method directly estimates latent vector $z$. However, the simplest way to estimate $z$ corresponding to each training data is to use an encoder network. In this experiment, we used a small encoder in addition to a pre-trained generator. The encoder and the scale and shift in the generator are updated during training, in which the generator is trained just like our method. The loss function is the sum of the L1 loss and perceptual loss.

\noindent{{\bf Ordinary training with proposed loss}}:
This method uses the same loss function as the proposed method, but updates all parameters or only a few layers of the generator instead of batch statistics. We tested three settings, updating only the first linear layer, updating the last residual block and any later layers, and updating all layers. These are called ``Update first", ``Update last", and ``Update all", respectively.

During these experiments, images were generated from 25 images sampled from each dataset. For the anime face dataset, we sampled images that have similar style features to limit the diversity of images. The details are provided in the supplementary material. We used the SNGAN projection model pre-trained on the ImageNet 1K dataset \cite{imagenet}. 

\begin{figure}[t]
\center
\includegraphics[width=1.0\linewidth]{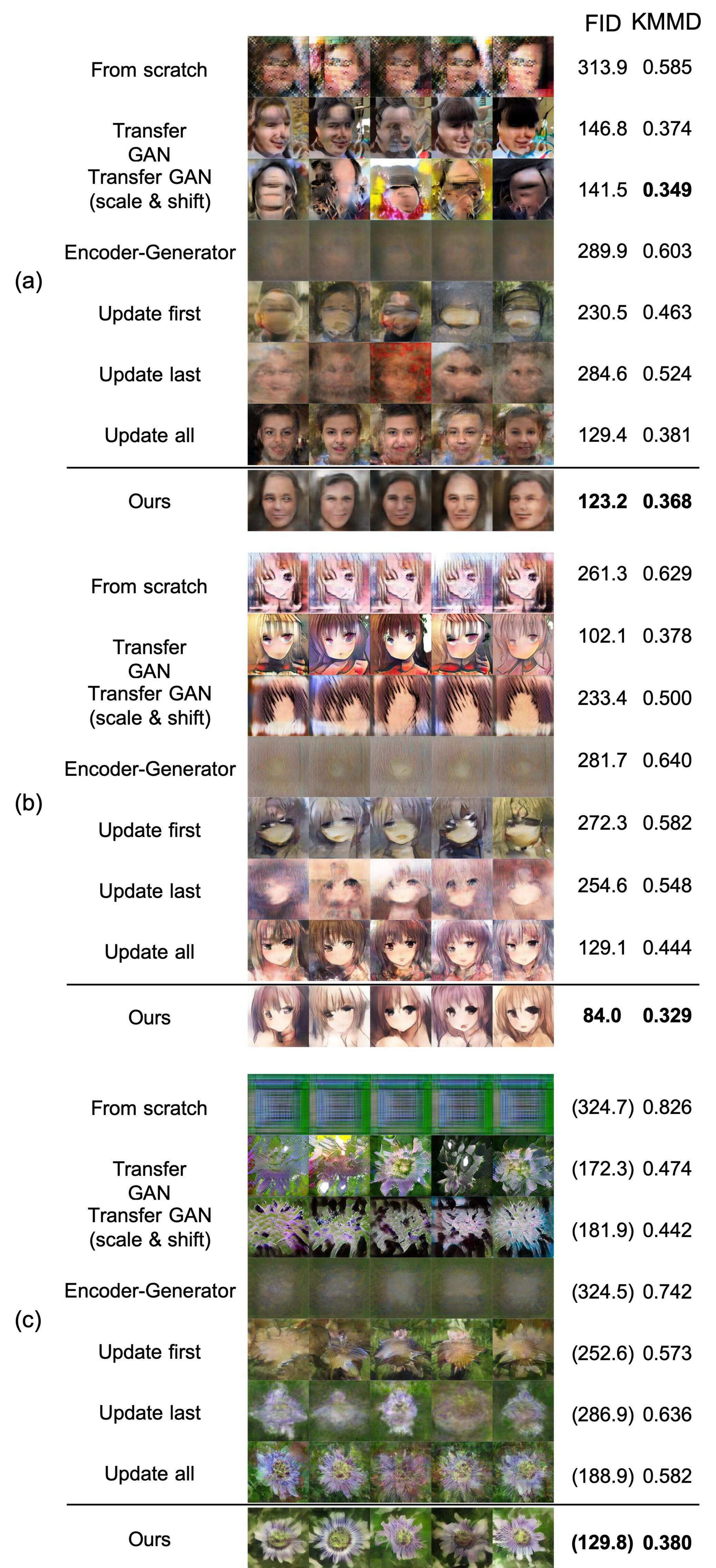}
\caption{Performance comparison on 25 training images for each method.}
\vspace{-5mm}
\label{other_results}
\end{figure}

To compare the training stability between our method and these methods, these models were trained for the same number of iterations as our method. For Transfer GAN, we stopped training before mode collapsing, and for ``Update all", we stopped training earlier because the model overfits early. Figure~\ref{other_results} shows random generative results and evaluation metrics from each method and dataset. 

GANs trained from scratch take time to converge, and blurry or meaningless objects can be generated. Transfer GAN converges more quickly, but the output is collapsed to a few modes. 
Transfer GAN (scale and shift) does not collapse but generates meaningless objects.
This means that the kernels of the pre-trained discriminator are optimized to estimate the distance between the generated and training images during pre-training, and is not suitable for estimating the distance of the distributions in the target domain. An encoder-generator takes time to converge, and blurred images are generated. 
Updating the first linear kernel can transfer global structure, but texture information cannot be transferred. Updating the last layers can only generate uncertain images because the model cannot acquire a global feature. Updating all layers overfits easily to training samples and can just generate intermediate color per pixel (See Figure~\ref{other_results} (c), Figure~\ref{flower_interpolation}, and the interpolation comparison in the supplementary materials).

These results show that adversarial training is unsuitable for image generation from a small dataset because the training is not stable. Besides, the supervised training tested in this experiment is also unsuitable for image generation from a small dataset because we have to train many parameters. Moreover, both global and local features must be transferred to generate target images, 
but updating all parameters easily overfit to training samples.
These results indicate the difficulty of this task.

\subsection{Proposed method on small dataset \label{generation_small}}
In this section, we conducted experiments with the proposed method and evaluated the performance. We used the same dataset and pre-trained model as used in the previous section.

\begin{figure}[t]
\center
\includegraphics[width=0.8\linewidth]{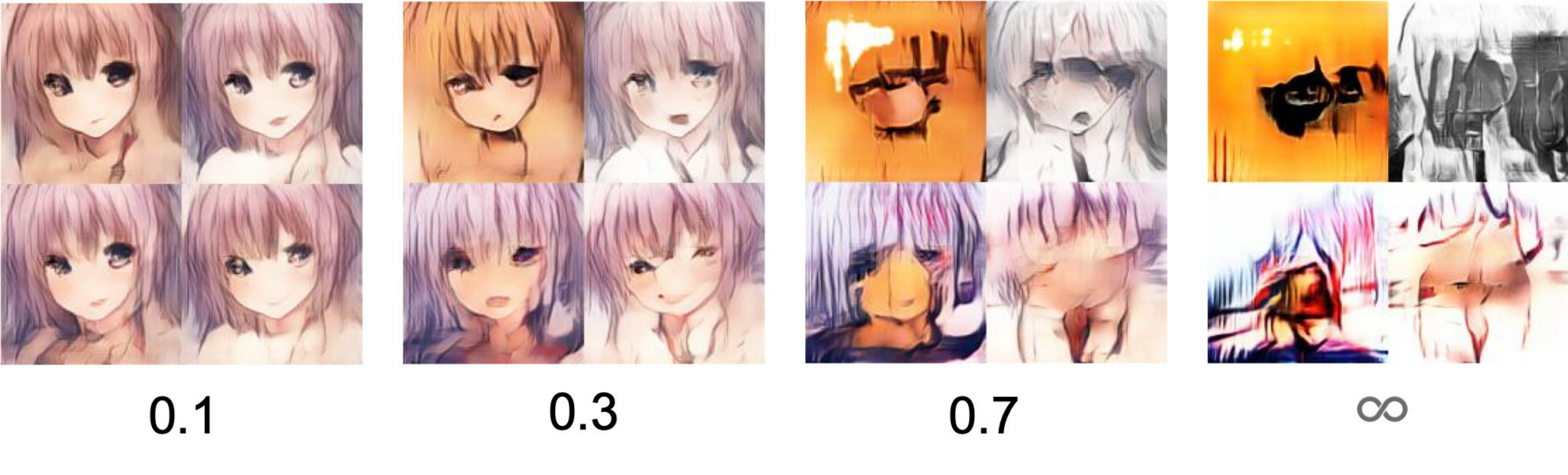}
\vspace{-3mm}
\caption{Sampled images from different truncated distributions. The number shown is the truncation threshold.}
\vspace{-3mm}
\label{normal}
\end{figure}

On the bottom row for each dataset in Figure~\ref {other_results} show the images generated from latent vectors $z$ sampled randomly from a truncated normal distribution. We used 0.3 or 0.4 for the truncation threshold. By truncating, it was confirmed that consistent images were generated. \figref {normal} shows images sampled from a normal distribution changing the truncation threshold. We confirmed that sampling with small truncation threshold can generate images with higher fidelity, and with larger threshold can generate images with diversity.

\begin{figure}[t]
\center
\includegraphics[width=1.0\linewidth]{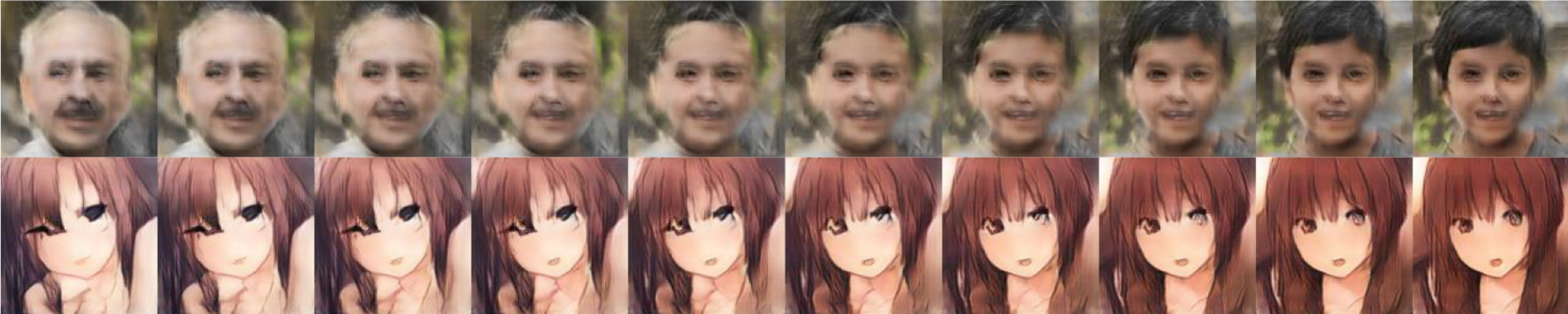}
\vspace{-3mm}
\caption{Interpolation between two generated images from adapted SNGAN on the human face dataset and anime face dataset containing 25 training samples.}
\label{sngan_interpolation}
\end{figure}

\begin{figure}[t]
\center
\includegraphics[width=1.0\linewidth]{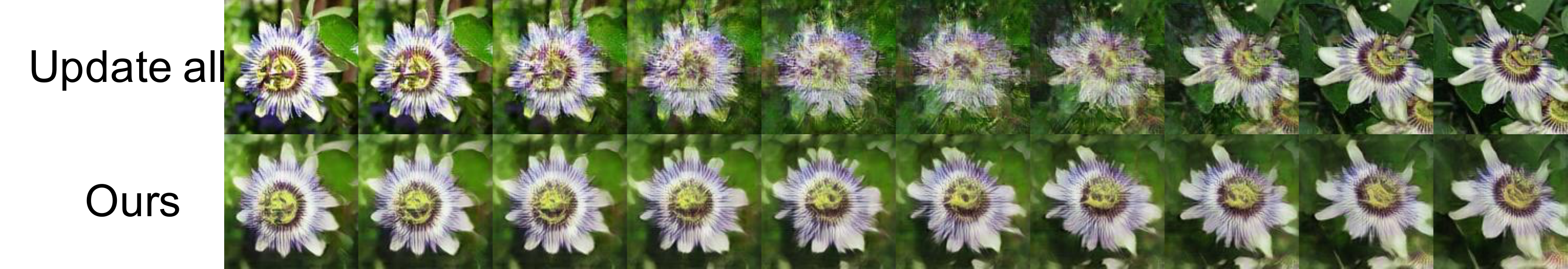}
\vspace{-3mm}
\caption{comparison of update\_all and ours on flower dataset. The proposed method performs more consistent interpolation.}
\label{flower_interpolation}
\end{figure}

From Figure~\ref {other_results}, we can see that the model converges stably without collapsing and can generate more consistent images. This is because the method effectively reuses the pretrained convolutional kernel parameters. 
Because the model only learns the relationship between latent vectors and sparse training samples, random generation is more difficult than interpolation.
Figure~\ref {sngan_interpolation} and Figure~\ref{flower_interpolation} shows the results of interpolation between two latent vectors $z$. Although the amount of training data is small, a smooth and consistent interpolation is achieved compared to other methods.

Comparing the performance with the other methods, our method can generate images with better FID and KMMD in general.
From this, it was shown that, even with a very small dataset such as 25 images, the proposed method can generate data with a semantic consistency in domains that other methods cannot.

\subsection{Dataset size and image quality}
In this section, we evaluate the relationship between the size of the dataset and the quality of the generated images.  Besides, we compared the performance of our method, Transfer GAN \cite{transfergan}, and ``Update all" when changing the size of the dataset. 

\begin{figure}[t]
\center
\includegraphics[width=1.0\linewidth]{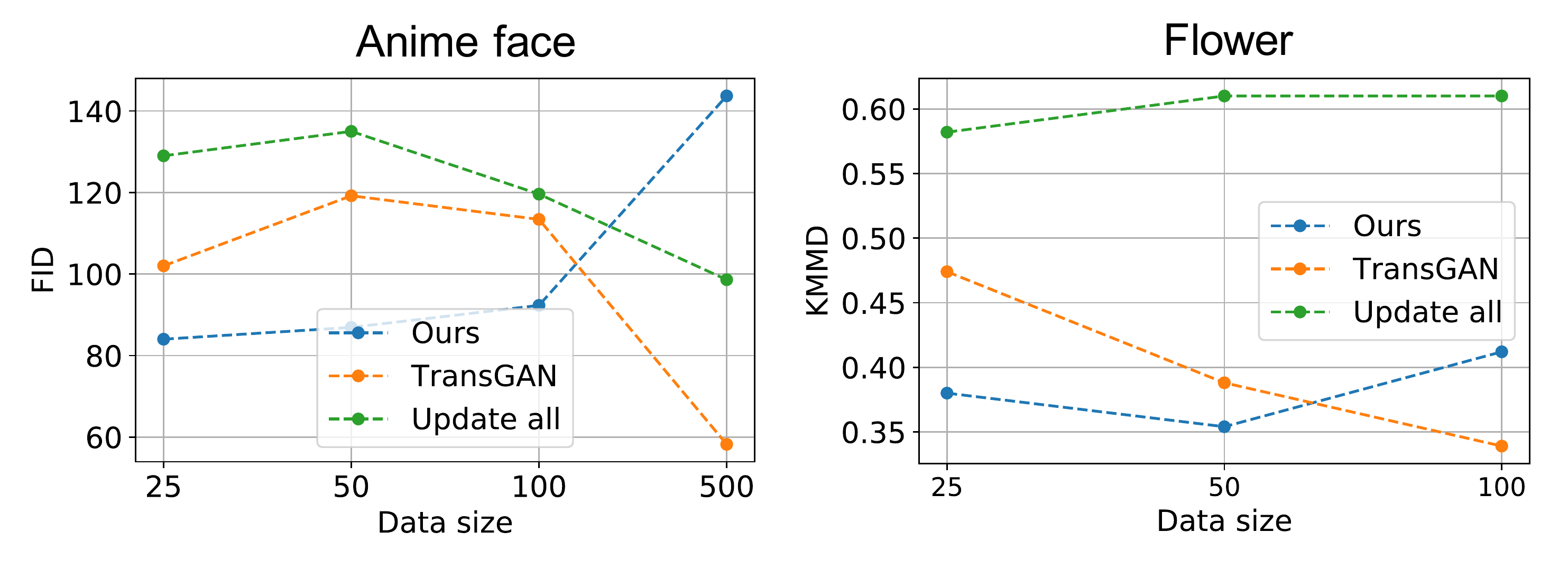}
\vspace{-3mm}
\caption{Comparison of FID on the anime face dataset (left) and KMMD on the flower dataset (right) between our method, Transfer GAN \cite{transfergan}, and ``Update all". Note that it is meaningless to compare the performance between different dataset sizes because the data distributions for each dataset size are different.}
\label{compare}
\end{figure}

The images generated randomly from the proposed models learned from each dataset of each data size are shown in Figure~\ref{datasize_face} and the generative results from TransferGAN and ``Update all" are shown in the supplementary materials. The scores for each model are shown in Figure~\ref{compare}. We report FID for the anime face dataset and KMMD for the flower dataset. The generated images shown in Figure~\ref{datasize_face} and in the supplementary materials show that when the size of the training dataset becomes large, blurred images compared to other methods tend to be generated. 
The evaluation scores in \figref{compare} show that for the anime face dataset, the proposed method works well compared to other methods when the data size is smaller than 500, and for the flower dataset, the proposed method generates better images compared to Transfer GAN when the data size is smaller than 100.  These results show that the quality of the generated images is limited for large datasets. On the other hand, the adversarial approach is very powerful when the dataset size increases. Also, ``updating all" performs better than our method when the datasize is large for the anime face dataset. This would be because it has higher degrees of freedom of trainable parameters.

As a result, though our model works well for small datasets, its performance is limited when the dataset size becomes large. This could be solved by combining adversarial approaches or increasing the trainable parameters.

\begin{figure}[t]
\center
\includegraphics[width=1.\linewidth]{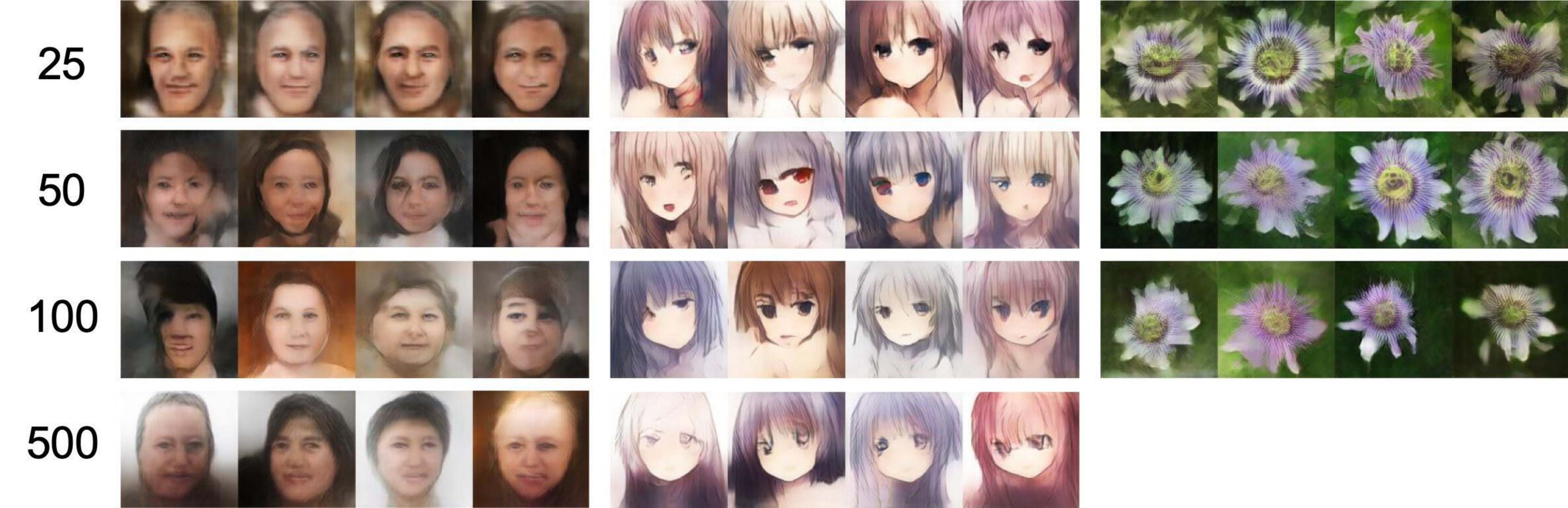}
\vspace{-3mm}
\caption{Randomly sampled human face, anime face, and flower images for each data size. The data sizes used for training are 25, 50, 100, and 500 from top to bottom for human face images (left), anime face images (middle), and flower images (right).}
\label{datasize_face}
\end{figure}

\subsection{Source domain selection}
In this subsection, we conducted an experiment to investigate when the generator can be transferred to the target domain. As discussed in \ref{scale_shift_analysis}, the diversity of the filter acquired in the pre-trained generator is thought to affect the quality of the generated images after the transfer. To investigate, we compared the transferred results to the anime face dataset and flower dataset adapted from a randomly initialized generator, a generator pre-trained on the face dataset \cite{celeba}, the LSUN bedroom dataset \cite{lsun}, and the ImageNet 1K. It can be inferred that a generator trained using the face and bedroom datasets has useful filters for image generation compared to a randomly initialized generator, although the diversity of the acquired filters is small compared to the generator trained using ImageNet.

\begin{figure}[t]
\center
\includegraphics[width=1.\linewidth]{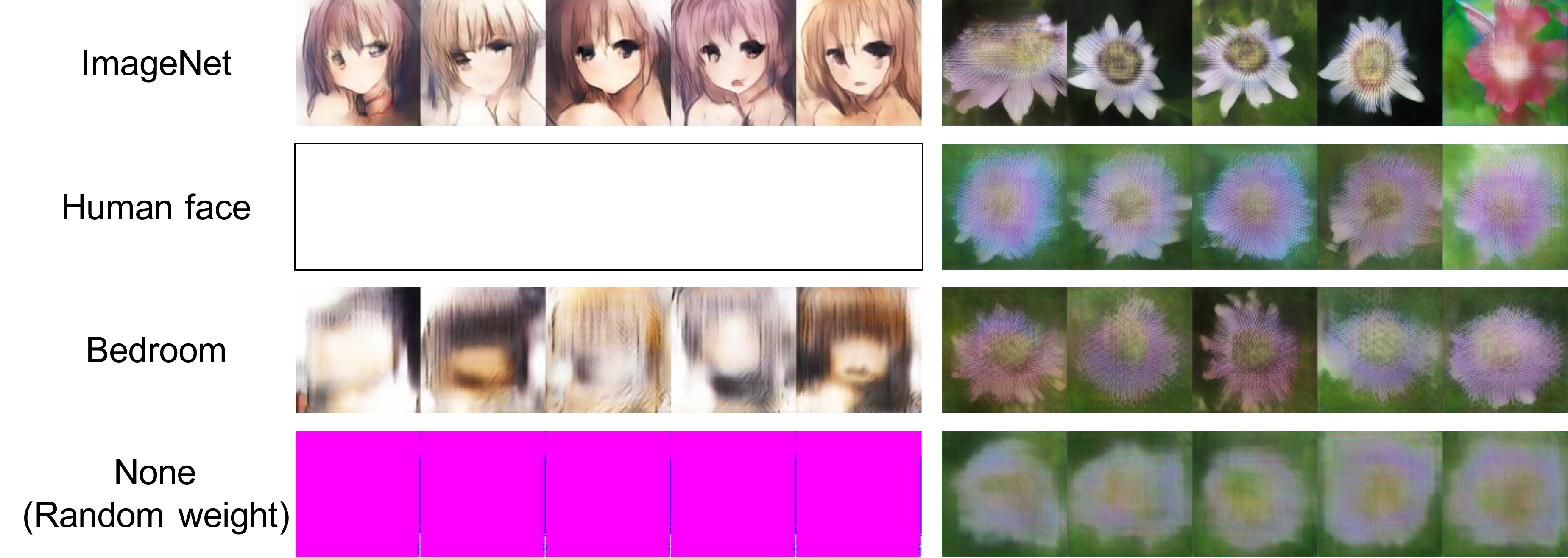}
\caption{Comparison of the image quality for different source datasets.}
\label{trans_comp}
\end{figure}

\figref{trans_comp} shows the results of four experimental settings.
As shown in the figure, the generator pre-trained on ImageNet 1K can generate images with the best quality. The generators transferred from other datasets generate blurry or meaningless images. This result shows that the diversity of the acquired filters in the source domain is important for adaptation. This does not limit our method as generators pre-trained on ImageNet are publically available.

\subsection{Higher resolution image synthesis}
\begin{figure}[t]
\center
\includegraphics[width=1\linewidth]{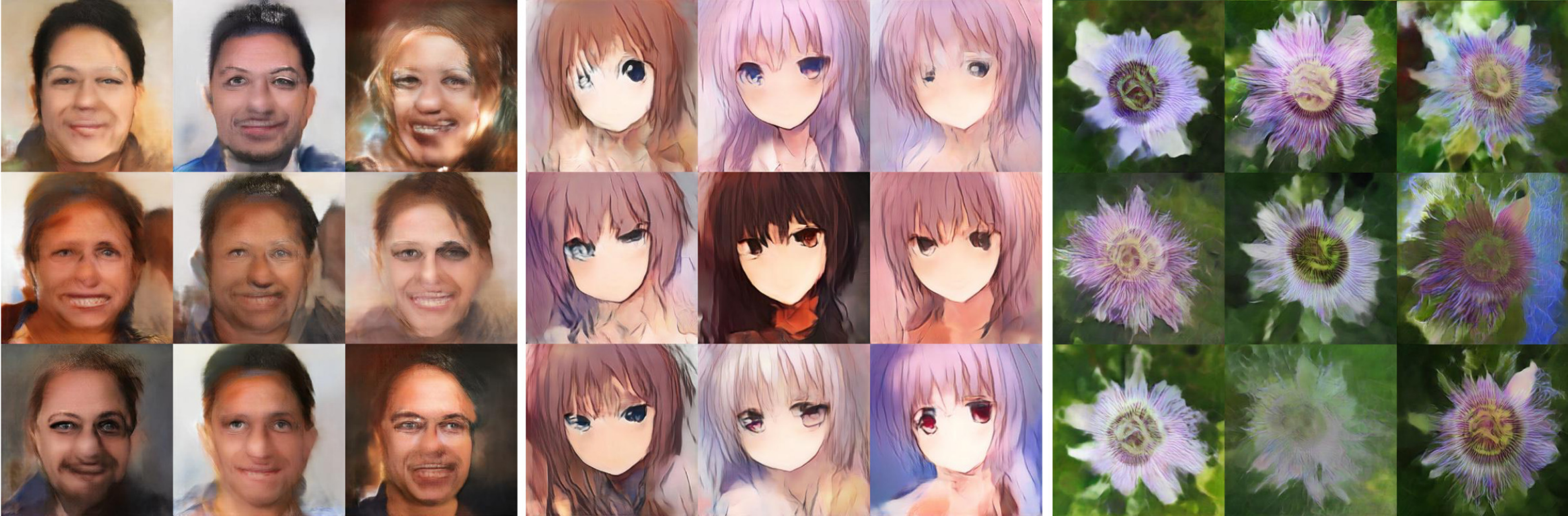}
\caption{Randomly sampled 256 $\times$ 256 images from adapted BigGAN on each dataset containing 25 training samples. The truncation threshold is 0.2.}
\label{biggan}
\end{figure}

\begin{figure}[t]
\center
\includegraphics[width=1\linewidth]{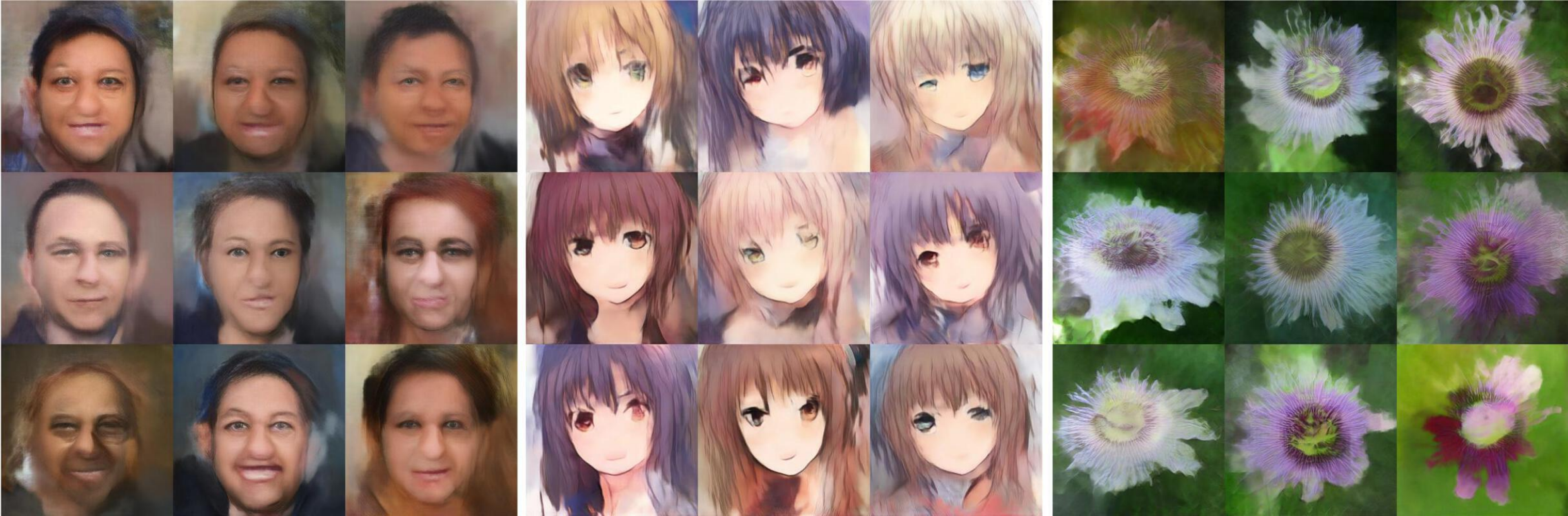}
\caption{Randomly sampled 256 $\times$ 256 images from adapted BigGAN on each dataset containing 50 training samples. The truncation threshold is 0.2.}
\label{biggan50}
\end{figure}

\begin{figure}[t]
\center
\includegraphics[width=1.0\linewidth]{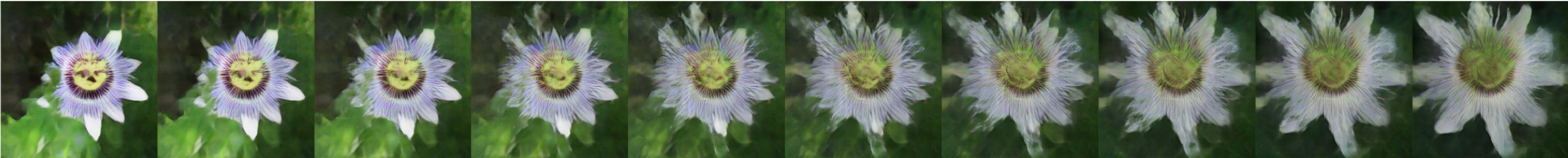}
\caption{Interpolation between two generated 256 $\times$ 256 images from adapted BigGAN on flower datasets containing 25 training samples.}
\label{biggan50_inter}
\end{figure}

We also applied our method to a pre-trained BigGAN-256 model on ImageNet. For conditional batch normalization, the parameters to calculate statistics were updated during training as they are calculated with neural nets. Because these neural nets have strong regularizations and are less likely to overfit, the last term in the \equref{loss} was not used for the statistics in the conditional batch normalization layer. We tested BigGAN on datasets consisting of 25 and 50 training samples. We found that updating the first linear kernel with a very small learning rate ($10^{-7}$) can generate sharper images for BigGAN.
The random generative results are shown in Figure~\ref{biggan} and \ref{biggan50}, and the interpolation results are shown in Figure~\ref{interpolation} and \ref{biggan50_inter}. Our method works well on higher resolution images.


\subsection{Domain addition to an existing generator}

The proposed method does not update the kernel parameters in the generator. Therefore, we can add new classes or domains without disturbing the performance on the source domain. This can be seen as low-shot learning \cite{lowshot} for a generator, and is a completely new task that is enabled only by our method.

\begin{figure}[t]
\center
\includegraphics[width=1.0\linewidth]{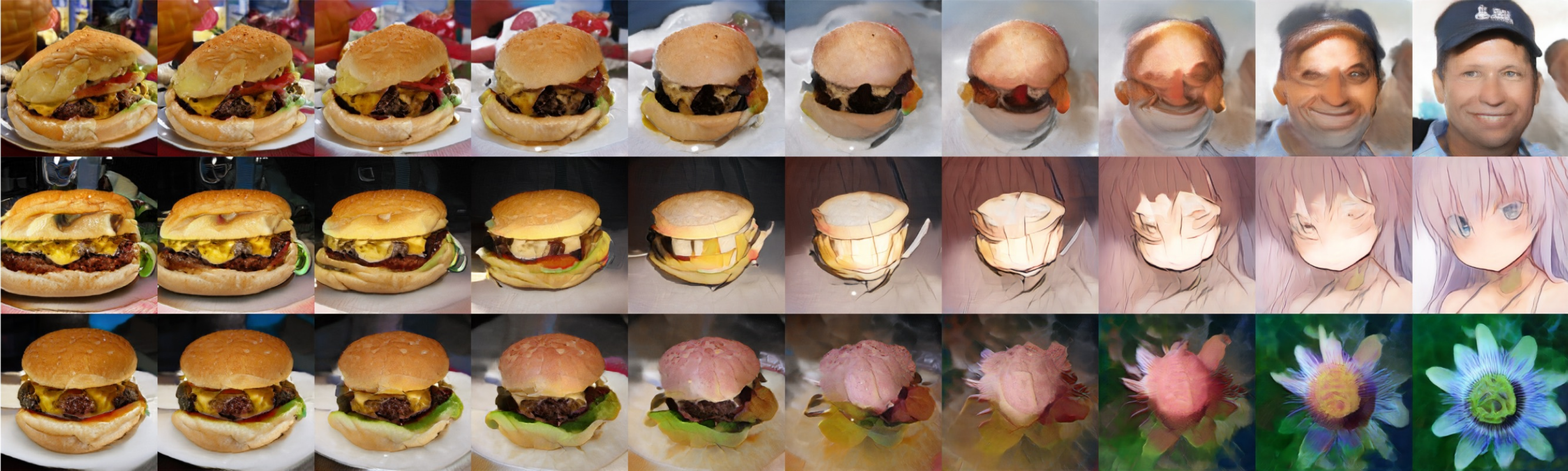}
\caption{Consecutive domain morphing, showing morphs from ``cheeseburger" in ImageNet to ``human face" in FFHQ dataset (top), from ``cheeseburger" to ``anime face" (middle), and from ``cheeseburger" to ``flower" (bottom).}
\label{morph}
\vspace{-4mm}
\end{figure}

From this, by changing the scale and shift parameters and latent vector, we can perform smooth morphing between different domains, similar to what is performed in \cite{sngan, biggan}. We conducted this domain morphing between ``cheeseburger" in ImageNet generated by the pre-trained generator and each target dataset. Smooth morphing is achieved, and the results are shown in \figref{morph}. This result shows that the generator uses the common knowledge for both source and target dataset.

\section{Conclusion}
In this work, we proposed a simple yet effective method for image generation from small datasets. By transferring the prior knowledge of a pre-trained generator and updating only the scale and shift parameters, it is possible to generate new images from much fewer images than required for regular generator training. Our results show that the proposed method can generate higher quality images using a small training dataset relative to existing methods.
This method can be used for a new task, low-shot learning for the generative model. As the proposed method can be leveraged for small dataset augmentation, future works include extending the method for classification tasks and few-shot learning.

\section{Acknowledgement}
This work was supported by JST CREST Grant Number JPMJCR1403, Japan. We would like to thank Antonio Tejero de Pablos, Audrey Chung, Hiroharu Kato, James Borg, Takuhiro Kaneko, and Yusuke Mukuta for helpful discussions.

\clearpage

{\small
\bibliographystyle{ieee_fullname}
\bibliography{egbib}
}

\clearpage

\renewcommand{\thefigure}{\Alph{figure}}
\renewcommand{\thetable}{\Alph{table}}
\setcounter{figure}{0}

\appendix

\twocolumn[
\maketitle\vspace*{54pt}
\begin{center}{\Large \bf Supplementary Material}\vspace*{54pt}\end{center}
]
\section{Experimental detail}

\subsection{Anime face sampling}
The Anime face dataset has a very high diversity in texture and facial shape and pose, which makes the dataset notably sparse when the dataset size is small. Therefore, we sampled images that have similar Gram matrix \cite{gram}, which is known to control the style information, to limit the textural diversity. This made the problem easier.

\subsection{Model selection}
We used SNGAN \cite{sngan} for unconditional GAN model, and 
SNGAN projection \cite{sngan, projection} and BigGAN \cite{biggan} for conditional GAN model. We used SNGAN-128 in the official SNGAN implementation\footnote{\textcolor[rgb]{1.0, 0.0, 1.0}{\url{github.com/pfnet-research/sngan_projection}}} and our reimplementation of BigGAN-256 with the official pretrained weight\footnote{\textcolor[rgb]{1.0, 0.0, 1.0}{\url{tfhub.dev/deepmind/biggan-256/2}}}. We used VGG16 \cite{vgg} trained on ImageNet for perceptual loss. We used layer ` conv1\_1', `conv1\_2', `conv2\_1', `conv2\_2', `conv3\_1', `conv3\_2', `conv3\_3', `conv4\_1', `conv4\_2', and `conv4\_3' for the perceptual loss. During training, for SNGAN, we updated scale and shift parameters of all conditional batch normalization layers and the fully connected layer in the generator. For BigGAN, statistics for the fully connected layer and all parameters to calculate batch statistics were updated.

\subsection{Training settings}
In this subsection, we describe the training setting for the experiments. Some experiments are trained with a different setting from the ones bellow. For details, please refer to our implementation.
\subsubsection{For SNGAN}
All models were trained for 3,000 to 4,000 training iterations with batchsize 25. We used Adam optimizer with initial learning rate 0.1 for datasize 25, 0.06 for datasize 50, 0.03 for datasize 100, and 0.02 for datasize 500. We used $\lambda_{C}=0.001$, $\lambda_z=0.2$, and $0\leq \lambda_{\gamma, \beta} \leq0.02$. For comparison experiments, we used Adam optimizer with learning rate 0.001 for encoders, and 0.0001 for the other GAN models. We used the L1 norm for the perceptual loss. It takes an hour on a single Nvidia P100 GPU for 3,000 training iterations.

\subsubsection{For BigGAN}
All models were trained for 6000 to 10000 training iterations with batchsize 16. We used Adam optimizer with learning rate 0.001 for the parameters of class embeddings, 0.0005 for other statistics parameters, and 0.05 for latent vectors. We used $\lambda_{C}^l=0.1/\sum_i \frac{1}{c_l h_l w_l}||C^{(l)}(x_i) - C^{(l)}(G(z_i + \epsilon))||_2$ instead of constant value, $\lambda_z=0.2$, and $\lambda_{\gamma, \beta} = 0$.  We found that with such $\lambda_C$, clearer images are generated. We used L2 norm for the perceptual loss. It takes 3 hours on 4 Nvidia P100 GPUs for 10,000 training iterations.

\section{Relationship between the scale and shift and activation rate}
\begin{figure}[t]
   \center
   \includegraphics[width=1.0\linewidth]{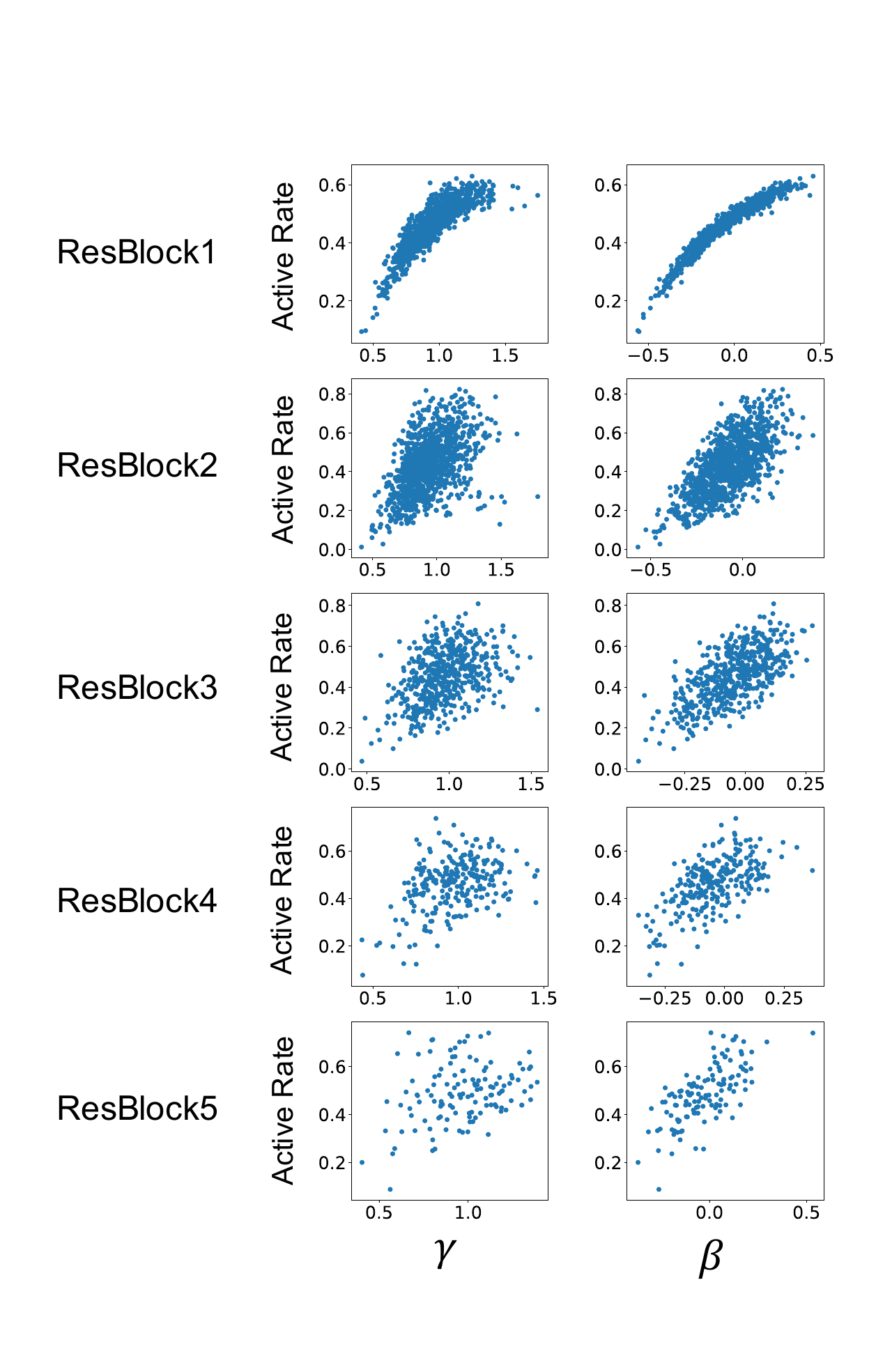}
   \vspace{-10mm}
   \caption{Relationship between the rate of active kernel and $\gamma$ and $\beta$.}
   \label{correlation}
\end{figure}

In Section 3, we stated that changing $\gamma$ and $\beta$ is equivalent to controlling the activation.
In this section, we investigated the relationship between the scale $\gamma$ and shift $\beta$ and the activation rate of each filter in SNGAN projection trained on ImageNet.
For the “blenheim spaniel” class, in \figref{correlation}, we show a plot of the relationship between $\gamma$ and $\beta$ during batch normalization and the activation rate of the output of the activation function in each layer, where each column indicates the results of the first conditional batch normalization in each residual block in the generator, where each point represents each filter. For all cases, a positive correlation exists between $\gamma$ and $\beta$ and the activation rate. Therefore, it can be stated that changing these parameters is equivalent to filter selection.

\section{Generated samples from Transfer GAN and ``Update all"}
\begin{figure}[t]
\center
   \includegraphics[width=1.0\linewidth]{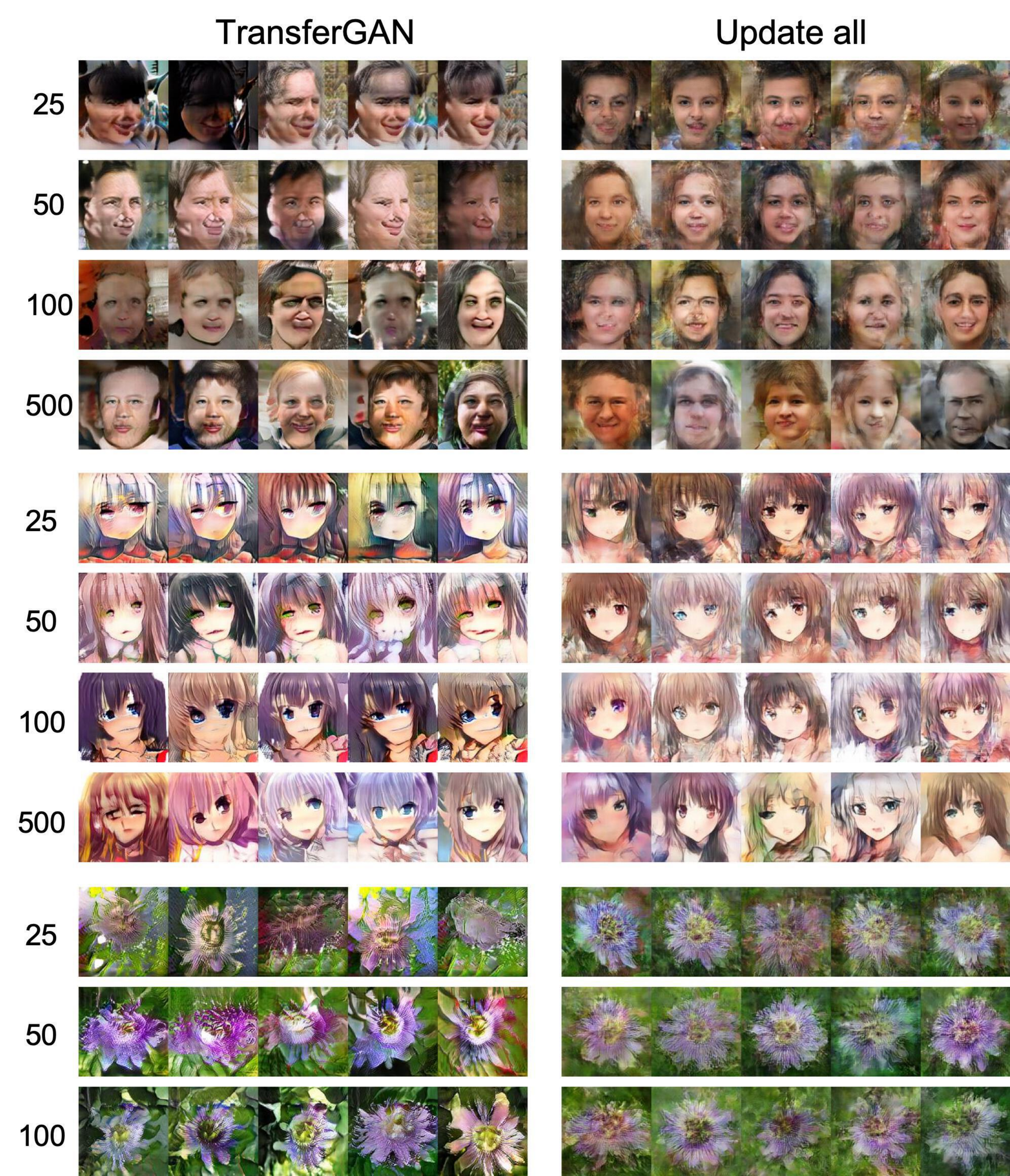}
   \vspace{-2mm}
   \caption{Generated images from Transfer GAN and ``Update all" trained with 25, 50, 100, 500 training images.}
   \label{trans25}
\end{figure}

In this section, we show the generative results from Transfer GAN \cite{transfergan} and ``Update all" for each data size, which transfers prior knowledge of generative models similarly to our method. We used human face, anime face, and flower images for training and chose 25, 50, 100, and 500 as the data size. We did not test 500 for flower dataset because the dataset has only 251 images. We stopped training the models before the generated images collapse for Transfer GAN. As seen in Figure~\ref{trans25}, when the dataset is small, the Transfer GAN generates similar images, though the model can generate clearer images than our method. ``Update all" just generate pixel-wise interpolation of training samples, which is apparent for flower dataset. This is discussed in the next section. Both methods generate images with better quality and diversity as the dataset sizes become large.

\section{Comparison of interpolation results}
\begin{figure}[t]
\center
   \includegraphics[width=1.0\linewidth]{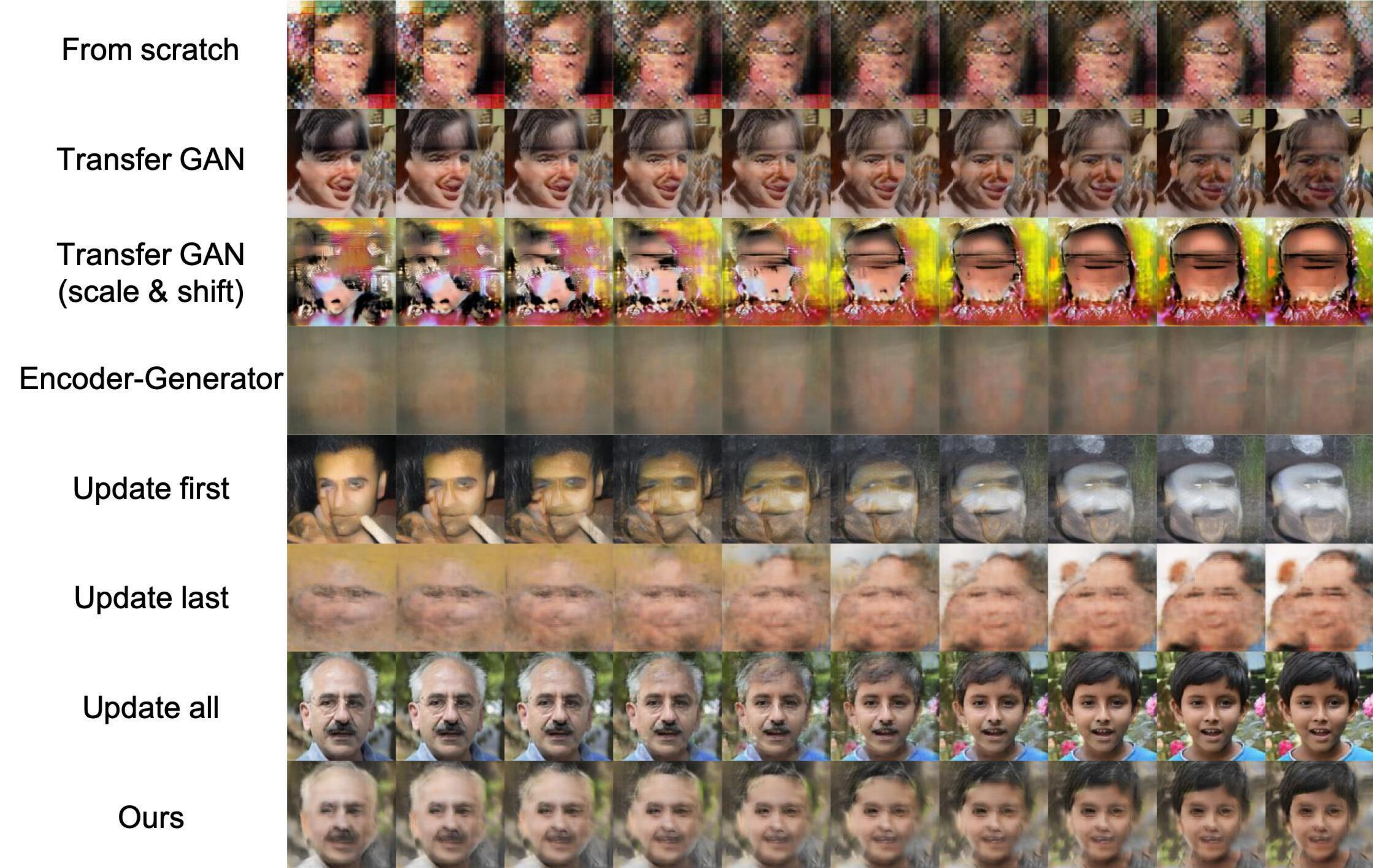}
   \vspace{-2mm}
   \caption{Generated human face images from all methods trained with 25 training images.}
   \label{inter_comp_ffhq}
\end{figure}

\begin{figure}[t]
\center
   \includegraphics[width=1.0\linewidth]{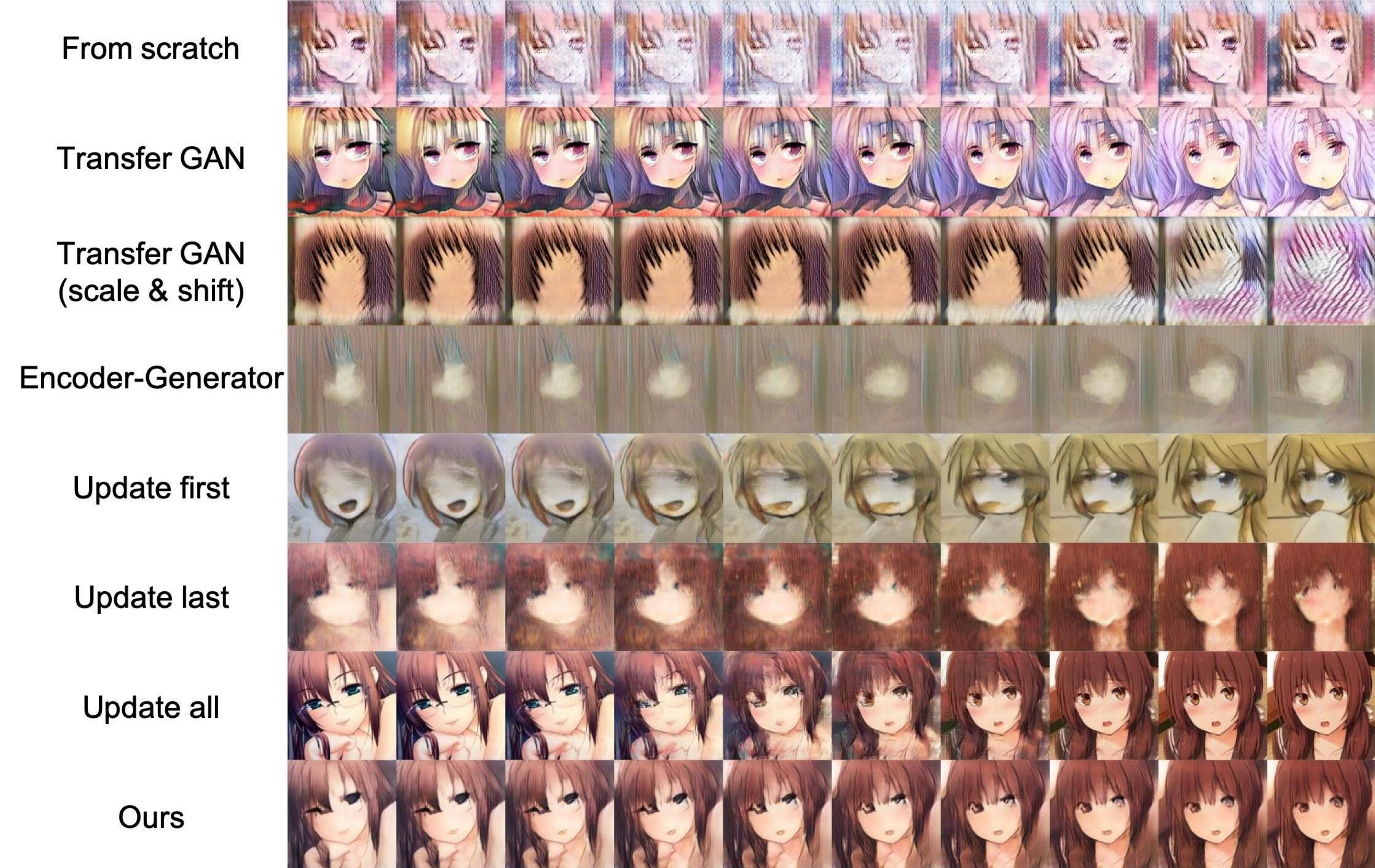}
   \vspace{-2mm}
   \caption{Generated anime face images from all methods trained with 25 training images.}
   \label{inter_comp_anime}
\end{figure}

\begin{figure}[t]
\center
   \includegraphics[width=1.0\linewidth]{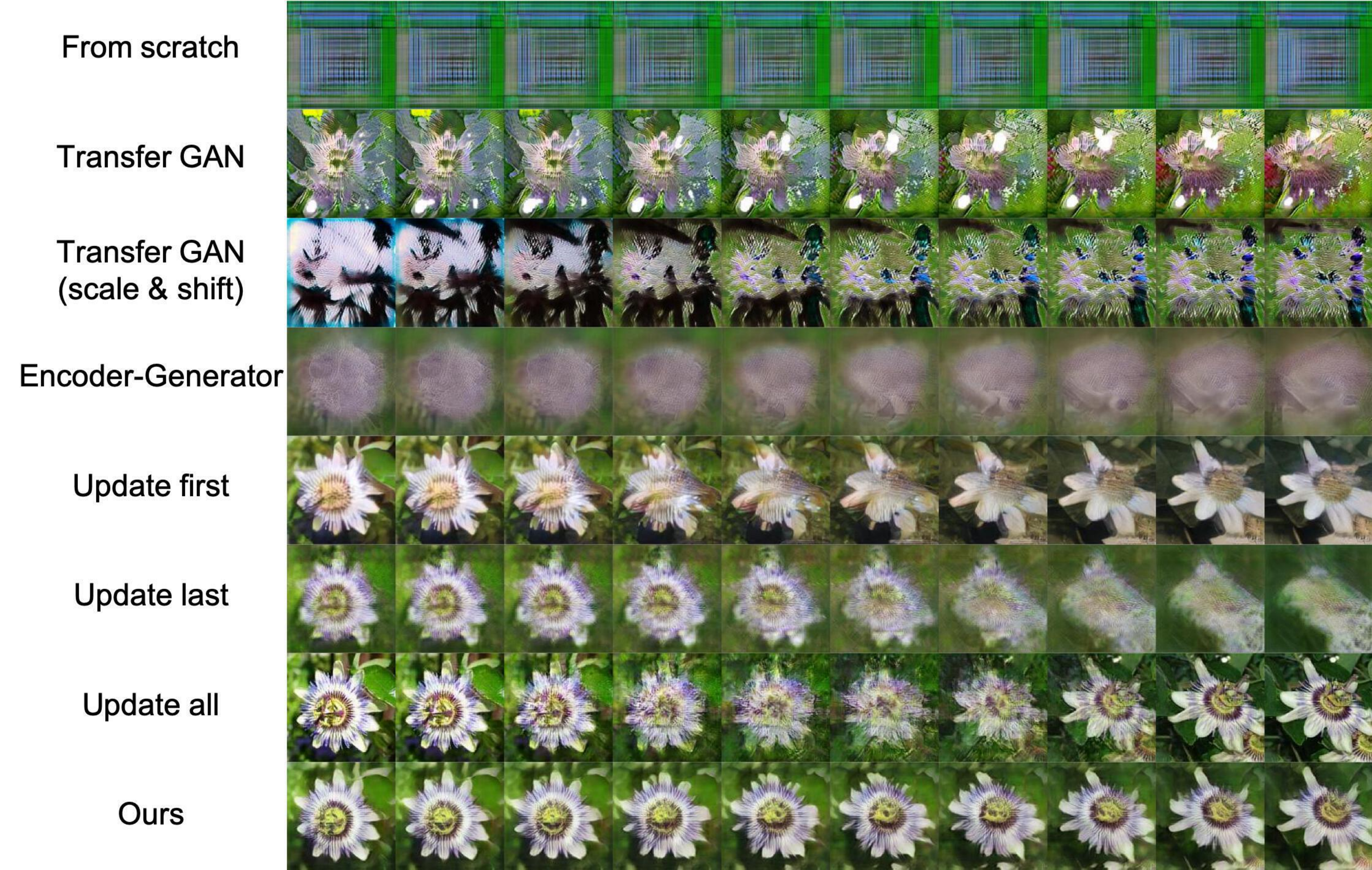}
   \vspace{-2mm}
   \caption{Generated flower images from all methods trained with 25 training images.}
   \label{inter_comp_flower}
\end{figure}

In this section, we show the interpolation results for each dataset when the models are trained on 25 training samples. In Figure~\ref{inter_comp_ffhq}, \ref{inter_comp_anime}, \ref{inter_comp_flower}, the top four rows show the interpolation between two randomly sampled images, and the bottom four rows show the interpolation between two generated images corresponding to two training samples.

The methods other than Transfer GAN, ``Update all", and ours generate images with limited quality. 
Transfer GAN seems to generate more consistent images but the generated images are collapsed to a few modes according to the random generation results and evaluation scores.
``Update all" just can conduct almost pixel-wise interpolation between two images. This is apparent for the hair change of human face and inconsistent shape of flower images.
On the other hand, our method can perform more consistent interpolation between two images, although they are a little blurry.

\end{document}